\documentclass[runningheads]{llncs}
\usepackage{amsfonts}
\usepackage{graphicx}
\usepackage{amsmath}
\usepackage[misc]{ifsym}
\usepackage{dcolumn}
\usepackage{amssymb}
\usepackage{etoolbox}
\usepackage{multirow}
\usepackage{booktabs}
\usepackage{todonotes}
\usepackage{hhline}
\usepackage{changes}
	\definechangesauthor[name={Percusse}, color=orange]{per}
\usepackage{array, makecell}
\usepackage{boldline}
\usepackage{colortbl}
\usepackage{paralist}

\usepackage{hyperref}
\hypersetup{
     colorlinks = true,
     citecolor = blue,
     linkcolor = blue,
     anchorcolor = blue,
     filecolor = blue,
     linkbordercolor = blue,
     urlcolor = blue
}
\usepackage{todonotes}

\begin{document}

\title{An Auto-Context Deformable Registration Network for Infant Brain MRI}
\titlerunning{AC-Reg-Net for Infant Brain MR Images}

\author{Dongming Wei\inst{1} \and Sahar Ahmad\inst{1} \and Yunzhi Huang\inst{1} \and Lei Ma\inst{1} \and Zhengwang Wu\inst{1} \and Gang Li\inst{1} \and Li Wang\inst{1} \and Qian Wang\inst{2}\textsuperscript{(\Letter)}  \and Pew-Thian~Yap\inst{1}\textsuperscript{(\Letter)} \and Dinggang Shen\inst{1}\textsuperscript{(\Letter)}}
\authorrunning{Anonymous}

\institute{Department of Radiology and Biomedical Research Imaging Center (BRIC), University of North Carolina, Chapel Hill, NC 27599, USA\\
\email{\{dgshen,ptyap\}@med.unc.edu}
\and Institute for Medical Imaging Technology, School of Biomedical Engineering,
Shanghai Jiao Tong University, Shanghai 200030, China\\ \email{wang.qian@sjtu.edu.cn}}
\maketitle              
\begin{abstract}
Deformable image registration is fundamental to longitudinal and population analysis. Geometric alignment of the infant brain MR images is challenging, owing to rapid changes in image appearance in association with brain development. In this paper, we propose an infant-dedicated deep registration network that uses the auto-context strategy to gradually refine the deformation fields to obtain highly accurate correspondences. Instead of training multiple registration networks, our method estimates the deformation fields by invoking a single network multiple times for iterative deformation refinement. The final deformation field is obtained by incremental composition of the deformation fields. Experimental results in comparison with state-of-the-art registration methods indicate that our method achieves higher accuracy while at the same time preserves the smoothness of the deformation fields. Our implementation is available \hyperlink{https://github.com/Barnonewdm/ACTA-Reg-Net}{online}.

\end{abstract}
\section{Introduction}
Deformable image registration~\cite{sotiras2013deformable,lester1999survey} establishes anatomical correspondences and is fundamental to longitudinal and population image analysis. 
Accurate registration of infant brain MRI is significantly more challenging than adults due to rapid shape and appearance changes of the brain images in association with dynamic development. In the first year of life, the overall brain volume doubles to about 65\% of the adult brain volume~\cite{knickmeyer2008a}. During this life span, gray matter (GM) develops more rapidly (108\% -- 149\%) than the white matter (WM) ($\sim$11\%), exhibiting significant increase in cortical thickness and surface area. 

Existing registration methods, such as SyN~\cite{avants2008symmetric}, diffeomorphic Demons~\cite{vercauteren2009diffeomorphic} and NiftyReg~\cite{rueckert1999nonrigid,modat2014global}, are based on iterative optimization and typically take a long time.
Recently, the deep learning based image registration methods~\cite{haskins2020deep} have been shown to predict deformations in a short time with high accuracy.
A registration network (Reg-Net) can be trained in an unsupervised manner with carefully designed metrics \cite{balakrishnan2019voxelmorph}.


It is however not straightforward to balance between image matching similarity and deformation smoothness. 
Balakrishnan~\textit{et al.}~\cite{balakrishnan2019voxelmorph} proposed to train the network with a loss function comprising of both image similarity and L2-norm of the deformation field. 
This method can get trapped in local optima while coping with the large deformation fields through a global regularization. 
Moreover, manual parameter fine-tuning is required in the training stage to decide the weight for loss term. To overcome these limitations, Dalca~\textit{et al.}~\cite{dalca2019unsupervised} implemented scaling and squaring based integration operations to estimate a velocity field, instead of directly predicting the deformation field. Hu~\textit{et al.}~\cite{hu2018adversarial} applied a discriminator network to gauge the smoothness of the output deformation. As the discriminator requires additional data for training, the applicability of the method can be limited. There are several other works~\cite{shen2019networks,zhao2019recursive} that train cascaded Reg-Nets to improve learning capacity, but at the expense of much more GPU memory due to the huge amount of parameters.

In this work, we propose an auto-context deformable registration network (AC-Reg-Net) for infant brain MRI. Instead of using cascaded training, AC-Reg-Net estimates the deformation fields by invoking a single network multiple times for incremental refinement of deformation fields. The final deformation field is obtained by composing all the incremental deformations. AC-Reg-Net functions very much in the spirit of auto-context modeling ~\cite{tu2010auto}.

\section{Methods}

\begin{figure}[t]
    \centering
    \includegraphics[width=\textwidth]{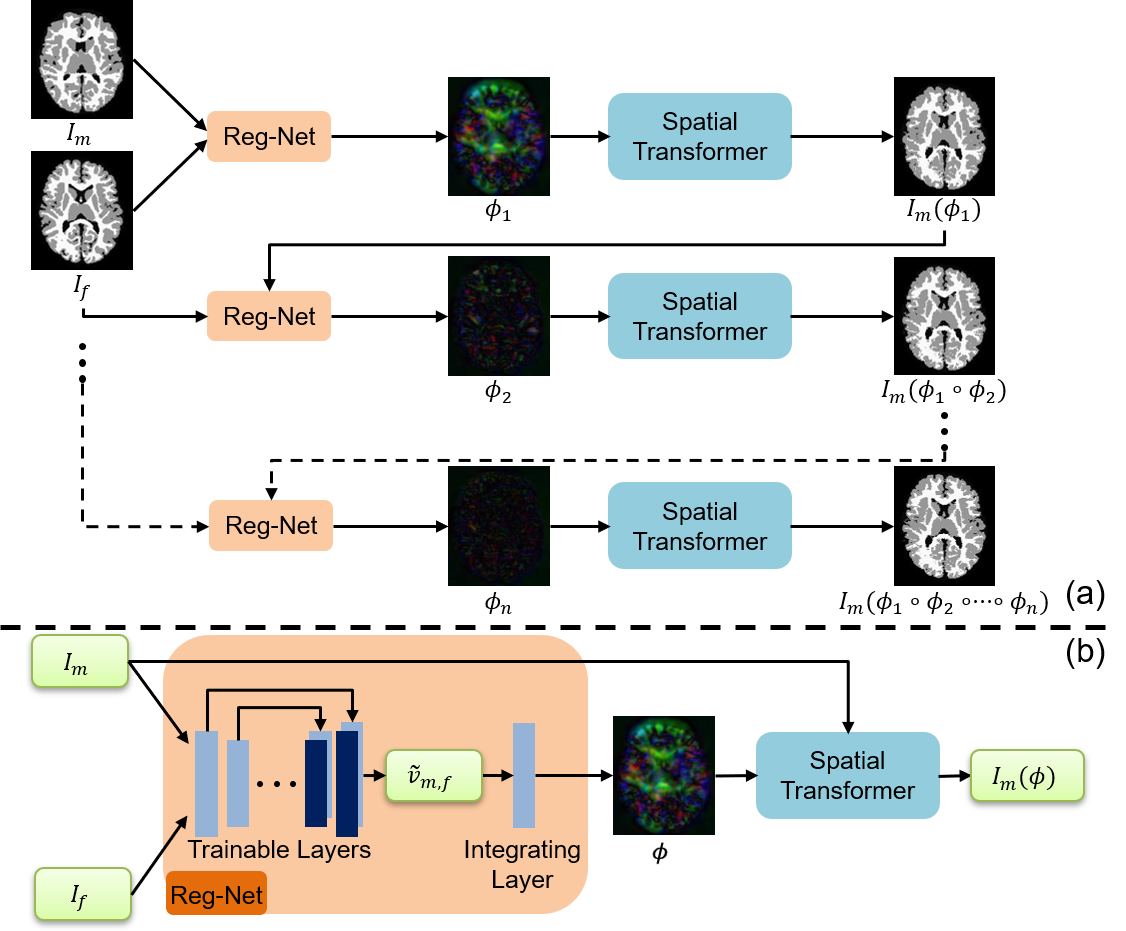}
    \caption{(a) Auto-context registration network (AC-Reg-Net). Deformation fields are color coded with red, green, and blue, representing deformation in the left-right, anterior-posterior, and inferior-superior directions, respectively. Black indicates zero displacement. (b) The architecture of the registration network (Reg-Net).} \label{fig1}
\end{figure}

AC-Reg-Net, illustrated in Fig.~\ref{fig1}, aims to progressively refine the deformation fields based on the `context' provided by prior estimates of the deformation. AC-Reg-Net consists of a basic deformable registration network (Reg-Net in the figure) and a spatial transformer~\cite{jaderberg2015spatial}.
The Reg-Net outputs smooth and incremental deformation fields. 
In our implementation, Reg-Net is trained following a tissue-aware topology-preserving metric based on tissue segmentation maps \cite{wang2019benchmark}. 
The spatial transformer, inspired by~\cite{jaderberg2015spatial}, resamples the moving tissue segmentation map based on the estimated deformation field. 
The Reg-Net and spatial transformer are invoked iteratively in a manner resembling auto-context modeling~\cite{tu2010auto}.



\subsection{Auto-Context Framework}
\label{Sect:auto-context}

The auto-context strategy (Fig.~\ref{fig1}(a)) consists of the following steps: 
\begin{enumerate}[(i)]
   \item The moving and fixed tissue segmentation map pair \{$I_{m}$, $I_{f}$\} is used as input to pre-trained Reg-Net to obtain the deformation field $\phi_{1}$ and the warped moving tissue segmentation map $I_{m}(\phi_{1})$.
   \item The new segmentation map pair \{$I_{m}(\phi_{1})$, $I_{f}$\} is fed into the same Reg-Net to get the new residual deformation field $\phi_{2}$ and the warped moving tissue segmentation map $I_{m}(\phi_{1}\circ\phi_{2})$. 
   \item The previous two steps are repeated $n$ times to obtain the final warped moving tissue segmentation map $I_{m}(\phi)$ with the final deformation field given by $\phi = \phi_{1}\circ\phi_{2}\circ...\circ\phi_{n}$.
\end{enumerate}
Note that, unlike~\cite{zhao2019recursive}, we avoid error accumulation of repeated segmentation map resampling by composing the deformation fields before warping the moving tissue segmentation map in each iteration.

\subsection{Deformable Registration Network}

The Reg-Net outputs a smooth deformation field, which when composed with prior deformation estimates, aims to lead to highly accurate and smooth alignment. 
Only a single pre-trained Reg-Net is utilized for all iterations. Our Reg-Net consists of trainable layers and an integration layer, as shown in Fig.~\ref{fig1}(b).

\subsubsection{Architecture --} The trainable layers are a 3D U-Net akin to VoxelMorph~\cite{balakrishnan2019voxelmorph}, where the dimensions of the input and output layers are adapted to the specific registration task. Given the moving/fixed tissue segmentation map pair \{$I_{m}, I_{f}$\}, the input layer size is $x\times y\times z \times2$ for two input segmentation maps, and the output layer size is $x\times y\times z \times3$ for the deformation field. Reg-Net estimates a velocity field $\Tilde{v}_{m,f}$ that can be integrated across iterations to eventually result in a smooth deformation field $\phi$, as inspired by LDDMM~\cite{beg2005computing}. 
The integrating layer adopts scaling and squaring operations as described~\cite{dalca2019unsupervised,ashburner2007a}. For training the Reg-Net in an unsupervised manner, the spatial transformer is applied to obtain the warped moving tissue segmentation map $I_{m}(\phi)$ (Fig.~\ref{fig1}(b)).

\subsubsection{Loss Function --} 
The loss function used to train Reg-Net consists of a dissimilarity function and a regularizer. Similar to LDDMM~\cite{beg2005computing}, the basic loss function is defined as
\begin{equation}
    \mathcal{L} = -\text{Sim}(I_{m}(\phi), I_{f}) + \mathrm{Reg}(\Tilde{v}_{m,f}),\;\;\phi= \int_{0}^{1}\Tilde{v}_{m,f},
\end{equation}
where Sim($\cdot$) can be any mono-modal similarity metric, which in our case is implemented as the localized normalized cross-correlation. Reg($\cdot$) is the L2-norm of the gradient of $\Tilde{v}_{m,f}$, as defined in VoxelMorph. However, the regularization of VoxelMorph is insufficient for avoiding folding in infant MRI registration, as we will demonstrate with experimental results. Therefore, we propose to regularize the deformation via tissue-aware Jacobian determinant $J(\phi)$ for greater smoothness. We constrain $J(\phi)$ to be positive via
\begin{equation}
            \mathrm{Reg}(\phi) = \mathrm{Reg}(J(\phi)) = \begin{cases}
            ||\mathrm{exp}(|\mathrm{min}(J) - 1|)-1||
            ; & \text{if}\;\text{GM or WM}\\
            ||\mathrm{exp}(|\mathrm{mean}(J) - 1|) - 1||; & \text{otherwise}.\\
            \end{cases}
\end{equation}
This regularization adaptively constraints the Jacobian determinant according to the tissues type. For GM and WM, the minimum of the Jacobian determinant should be positive. For the background and CSF, the average Jacobian determinant should be close to 1. 
The loss function is defined as
\begin{equation}
    \mathcal{L} = -\text{Sim}(I_{m}(\phi), I_{f}) + \mathrm{Reg}(\Tilde{v}_{m,f}) + \mathrm{Reg}(\phi),\;\;\phi= \int_{0}^{1}\Tilde{v}_{m,f}.
\end{equation}

\subsubsection{Implementation --} The proposed Reg-Net was implemented in Keras and trained on a single 12GB NVIDIA Titan X GPU. We used ADAM optimizer with a learning rate of $1~\times~10^{-4}$. The network was trained for 1500 epochs, with 100 iterations in each epoch. The Reg-Net was trained by randomly selecting one tissue segmentation map pair from the training dataset in each iteration, and it took around 145 hours to finish the entire training processing.

\section{Results and Discussion}
\subsubsection{Dataset and Preprocessing --} 
The dataset consisted of longitudinal T1w and T2w images (acquired at 2 weeks, 3, 6 and 12 months after birth) of 47 healthy infant subjects enrolled as part of the \textit{anonymous} study. The imaging parameters for T1w MR images were: TR = 1900 ms, TE = 4.38 ms, flip angle = 7$^{\circ}$, 144 sagittal slices, and 1 mm isotropic voxel resolution. The imaging parameters for T2w MR images were TR = 7380 ms, TE = 119 ms, flip angle = $150^{\circ}$, 64 sagittal slices, and $1.25\times1.25\times1.95\,\text{mm}^3$ voxels resolution. The dataset is pre-processed by the infant dedicated pre-processing pipeline~\cite{sled1998a} 
to obtain tissue segmentation maps.

The number of scans for each subject can vary due to missed scans. The training dataset comprised of 56 longitudinal scans of 29 subjects, and 57 scans of the remaining 18 subjects were used for testing. We selected a 12-month-old tissue segmentation map from the testing dataset as the fixed image in the testing stage. All the tissue segmentation maps were rigidly aligned with the fixed tissue segmentation map using FLIRT~\cite{jenkinson2001a}. All the tissue segmentation maps and intensity images were then resampled to have a size of $256\times256\times256$ with $1\times1\times1$ mm$^3$ voxels resolution.

    \begin{figure}[t]
        \centering
        \includegraphics[width=0.9\textwidth]{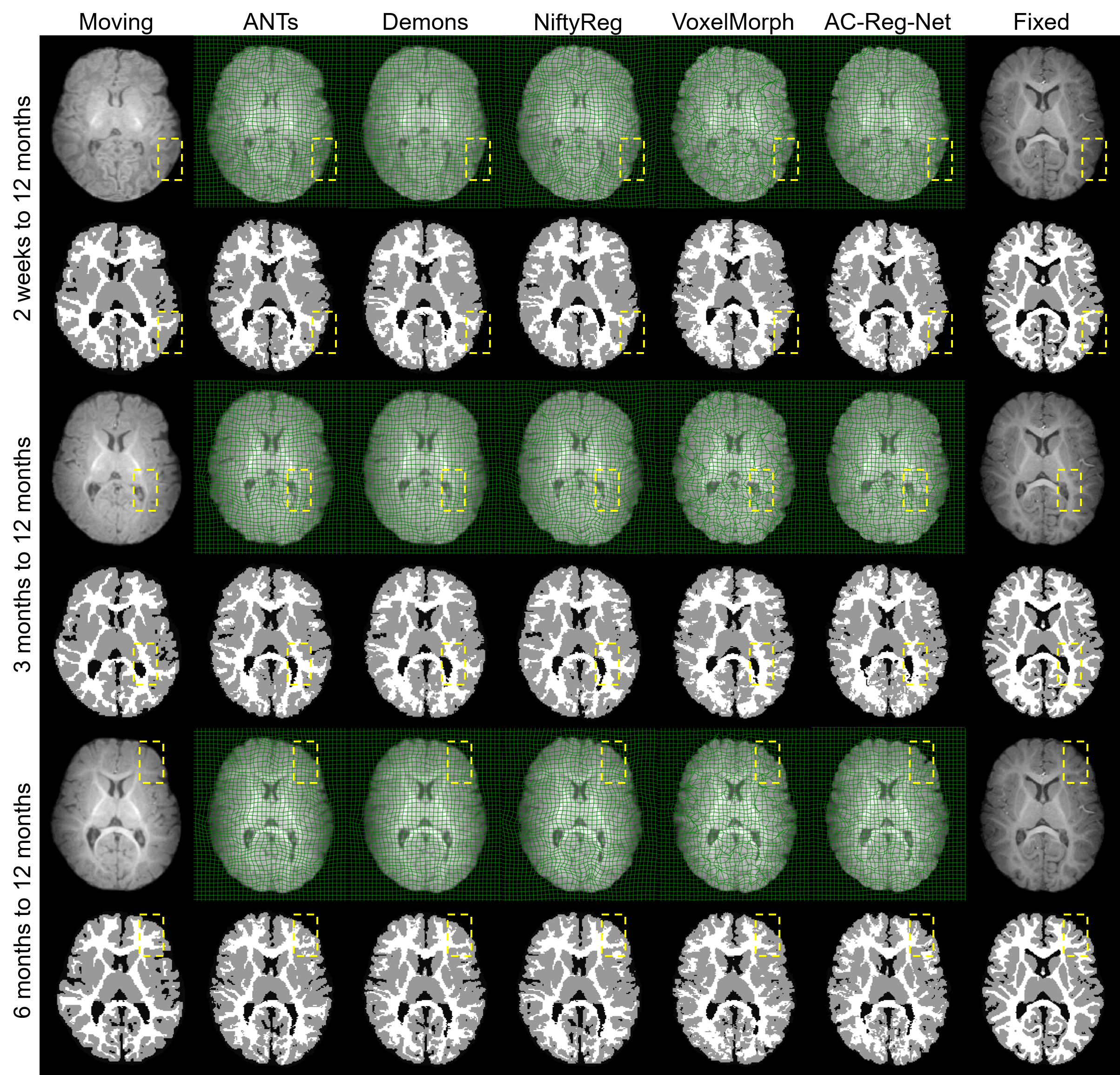}
        \caption{Results obtained with various registration methods, including warped intensity images overlaid with deformation fields, and warped tissue segmentation maps.}
        \label{fig:result}
    \end{figure}

\subsubsection{Evaluation Metrics --} We computed Dice similarity coefficient (DSC) over the segmented GM and WM. The smoothness of the deformation field was evaluated using ratio of folding points (RFP), which is the ratio of the negative Jacobian determinant voxels to the total number of voxels. 
Higher DSC with smaller RFP signifies better performance, meaning higher similarity with a more regularized deformation field. 
    
    \begin{figure}[t]
        \centering
        \includegraphics[width=\textwidth]{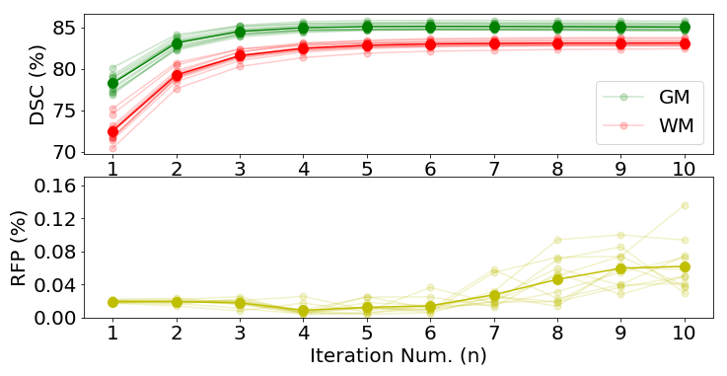}
        \caption{The mean DSC of GM and WM over the testing dataset (\textit{top}), and RFP of the deformation field (\textit{bottom}) with different numbers of iterations in the auto-context framework.}\label{fig:iteration}
    \end{figure}

\subsection{Comparison with Existing Methods}
We performed inter-subject registration over the testing dataset and randomly selected a 12-month-old scan as the fixed image. All the other segmentation maps in the testing dataset were then registered to this fixed image. We compared our method with SyN in the ANTs toolkit~\cite{avants2008symmetric}, diffeomorphic Demons~\cite{vercauteren2009diffeomorphic}, NifityReg~\cite{rueckert1999nonrigid}, and VoxelMorph~\cite{dalca2019unsupervised}. The parameters details for ANTs, Demons, and NifityReg are as follows: 
\begin{enumerate}[(1)]
   \item \hyperlink{http://stnava.github.io/ANTs/}{ANTs}: 
    
        \textbf{ANTS} 3 -m PR[fixed\_image.nii, moving\_image.nii, 1, 2] -O $<$output$>$ -i $30\times99\times11$ -t SyN[0.5] -r Gauss[2,0] --continue-affine false --use-NN -G
        
        \textbf{WarpImageMultiTransform} 3 moving\_labels.nii output\_labels.nii -R fixed \_image.nii outputWarp.nii outputAffine.txt -–use-NN
   \item \hyperlink{https://www.insight-journal.org/browse/publication/154}{Demons}: 
    
        \textbf{DemonsRegistration} -f fixed\_image.nii -m moving\_image.nii -O output.mha -e -s 2 -i 30x20x10 
        
        \textbf{DemonsWarp} -m moving\_labels.nii -b output.mha -o output\_labels.nii -I
   \item \hyperlink{https://cmiclab.cs.ucl.ac.uk/mmodat/niftyreg}{NiftyReg}:
    
        \textbf{ref\_f3d} -flo moving\_image.nii -ref fixed\_image.nii -res warped.nii -cpp output.nii
        
        \textbf{reg\_resample} -ref fixed\_image.nii -flo moving\_labels.nii -res output\_labels.nii -trans output.nii -inter 0
\end{enumerate}
We trained VoxelMorph with the training dataset using its default parameters~\cite{dalca2019unsupervised}. The quantitative results for the registration of 2 weeks to 12 months, 3 months to 12 months, and 6 months to 12 months are given in Table~\ref{tab:GM_WM_Dice}. It can be observed that AC-Reg-Net obtained significant improvement for DSC over the compared methods for all three time-points. The results can be visually inspected in Fig.~\ref{fig:result}, confirming that AC-Reg-Net obtains accurate alignment with smooth deformation fields compared to the other methods.

    \begin{table*}[h]
        \centering
        \caption{DSC (\%) and RFP (\%) over GM and WM by FLIRT, diffeomorphic Demons, ANTs, NiftyReg, VoxelMorph, Reg-Net, and AC-Reg-Net.}
        \resizebox{\textwidth}{!}{\begin{tabular}{|c|c|c|c|c|c|c|c|c|}
        \hline
                & \multicolumn{2}{c|}{\textbf{2 weeks to 12 months}} & \multicolumn{2}{c|}{\textbf{3 months to 12 months}} & \multicolumn{2}{c|}{\textbf{6 months to 12 months}} & \multirow{3}{*}{\textbf{RFP (\%)}}\\ \cline{2-7}
                   & \multicolumn{2}{c|}{DSC} & \multicolumn{2}{c|}{DSC} & \multicolumn{2}{c|}{DSC} & \\ \cline{2-7}
                   & GM & WM & GM & WM & GM & WM &\\ \hline
             FLIRT & $59.94\mathrm{\pm}1.64$ & $64.28\mathrm{\pm}0.91$ & $61.15\mathrm{\pm}1.32$ & $53.61\mathrm{\pm}1.16$ & $61.85\mathrm{\pm}1.47$ & $56.78\mathrm{\pm}1.40$ & \textbf{0}\\
             ANTs & $79.26\mathrm{\pm}0.84$ & $75.45\mathrm{\pm}1.01$ & $79.75\mathrm{\pm}0.67$ & $75.68\mathrm{\pm}1.14$ & $80.57\mathrm{\pm}0.61$ & $77.73\mathrm{\pm}0.73$ & 0.0453\\
             Demons & $73.60\mathrm{\pm}0.87$ & $68.99\mathrm{\pm}0.73$ & $74.03\mathrm{\pm}0.68$ & $69.01\mathrm{\pm}0.84$ & $74.70\mathrm{\pm}0.67$ & $71.52\mathrm{\pm}0.71$ & 0.0006\\
             NiftyReg & $72.65\mathrm{\pm}1.56$ & $68.29\mathrm{\pm}1.60$ & $73.61\mathrm{\pm}1.23$ & $68.70\mathrm{\pm}0.91$ & $74.56\mathrm{\pm}0.70$ & $72.94\mathrm{\pm}3.98$ & 0.0034\\
             VoxelMorph & $77.23\mathrm{\pm}6.60$ & $72.74\mathrm{\pm}6.70$ & $79.89\mathrm{\pm}4.75$ & $75.78\mathrm{\pm}4.65$ & $80.01\mathrm{\pm}5.53$ & $76.34\mathrm{\pm}6.22$ & 0.6915\\ \hline
             Reg-Net & 77.62${\pm}$0.96 & $71.69{\pm}0.85$ & $78.21{\pm}0.57$ & $71.99{\pm}0.51$ & $79.12{\pm}0.74$ & $74.27{\pm}0.87$ & 0.0193\\ 
             AC-Reg-Net & \textbf{84.96$\boldsymbol{\pm}$0.49} & \textbf{82.58$\boldsymbol{\pm}$0.56} & \textbf{85.12$\boldsymbol{\pm}$0.36} & \textbf{82.79$\boldsymbol{\pm}$0.18} & \textbf{85.17$\boldsymbol{\pm}$0.31} & \textbf{83.19$\boldsymbol{\pm}$0.10} & 0.0122\\
        \hline
        \end{tabular}}
        \label{tab:GM_WM_Dice}
    \end{table*}

\subsection{Benefits of Auto-Context Registration}
We evaluated the effects of the number of iteration in the auto-context framework. 
We compared the mean WM and GM DSC and RFP over the testing dataset.
Fig.~\ref{fig:iteration} shows that result varies with the iteration number. 
The mean DSC of GM and WM improves sharply from the 1st to the 2nd iteration and plateaus after the 5th iteration. RFP is kept very low for all the iterations. 
Since the deformation field $\phi_{i}$ generated by AC-Reg-Net is smooth for each iteration, the composition of these smooth deformation fields results in a smooth final deformation field $\phi$. We chose $n=5$ for AC-Reg-Net since the performance is optimal at this point.

\section{Conclusion}
This paper presented a deep registration framework for infant brain MRI. 
To counter appearance changes, our method, AC-Reg-Net, uses tissue segmentation maps for training.
AC-Reg-Net is applied in an auto-context manner, leveraging context information iteratively to improve registration accuracy. 
The Jacobian regularizer constrained the estimated deformation fields so that they are topology-preserving.
Experimental results validate the efficacy of AC-Reg-Net in registering MRI scans of infants of 2-week, 3-months, and 6-months of age to a 12-month-old scan, both in terms of accuracy and deformation regularity.

\bibliographystyle{splncs04}
\bibliography{bibex.bib}

\end{document}